\documentclass[journal]{IEEEtran}
\usepackage{graphicx}  
\usepackage{algorithm}
\usepackage{booktabs} 
\usepackage{algorithmic}
\usepackage{graphicx}
\usepackage{multirow}

\usepackage{times}
\usepackage{epsfig}
\usepackage{graphicx}
\usepackage{amsmath}
\usepackage{amssymb}
\usepackage{graphicx}  
\usepackage{algorithm}
\usepackage{booktabs} 
\usepackage{algorithmic}
\usepackage{graphicx}
\usepackage{multirow}
\usepackage{color}

\ifCLASSINFOpdf
\else
\fi
\begin{document}
%
\title{Automatic Salient Object Detection for Panoramic Images Using Region Growing and Fixation Prediction Model}
%
%
%

\author{Chunbiao~Zhu,~\IEEEmembership{Student~Member,~IEEE,}
        Kan~Huang,~\IEEEmembership{}
        Thomas H.~Li,~\IEEEmembership{}
        Ge~Li,~\IEEEmembership{Member,~IEEE,}
\thanks{The authors are with the School of Electronic and Computer Engineering,
Peking University Shenzhen Graduate School, Shenzhen 518055, China
(e-mail: zhuchunbiao@pku.edu.cn; khuang@pkusz.edu.cn; thomas.li@gpower-semi.com; gli@pkusz.edu.cn).}
}

%
%

\markboth{Journal of \LaTeX\ Class Files,~Vol.~14, No.~8, August~2015}%
{Shell \MakeLowercase{\textit{et al.}}: Bare Demo of IEEEtran.cls for IEEE Journals}
%



\maketitle

\begin{abstract}
Almost all previous works on saliency detection have been dedicated to conventional images, however, with the outbreak of panoramic images due to the rapid development of VR or AR technology, it is becoming more challenging, meanwhile valuable for extracting salient contents in panoramic images.

In this paper, we propose a novel bottom-up salient object detection framework for panoramic images. First, we employ a spatial density estimation method to roughly extract object proposal regions, with the help of region growing algorithm. Meanwhile, an eye fixation model is utilized to predict visually attractive parts in the image from the perspective of the human visual search mechanism. Then, the previous results are combined by the maxima normalization to get the coarse saliency map. Finally, a refinement step based on geodesic distance is utilized for post-processing to derive the final saliency map.

To fairly evaluate the performance of the proposed approach, we propose a high-quality dataset of panoramic images (SalPan). Extensive evaluations demonstrate the effectiveness of our proposed method on panoramic images and the superiority of the proposed method against other methods.
\end{abstract}

\begin{IEEEkeywords}
Saliency detection, eye fixation, region growing, panoramic images.
\end{IEEEkeywords}

%
\IEEEpeerreviewmaketitle

\section{Introduction}
%
%
%
%
\IEEEPARstart{A}{n} inherent and powerful ability of human eyes is to
quickly capture the most conspicuous regions from a scene, and passes them
to high-level visual cortexes. The neurobiological attention mechanism reduces the complexity
of visual analysis and thus makes human visual system considerably efficient
in complex scenes.

\begin{figure}[!h]

\begin{center}
\includegraphics[width=3.2in,height =1.4884in]{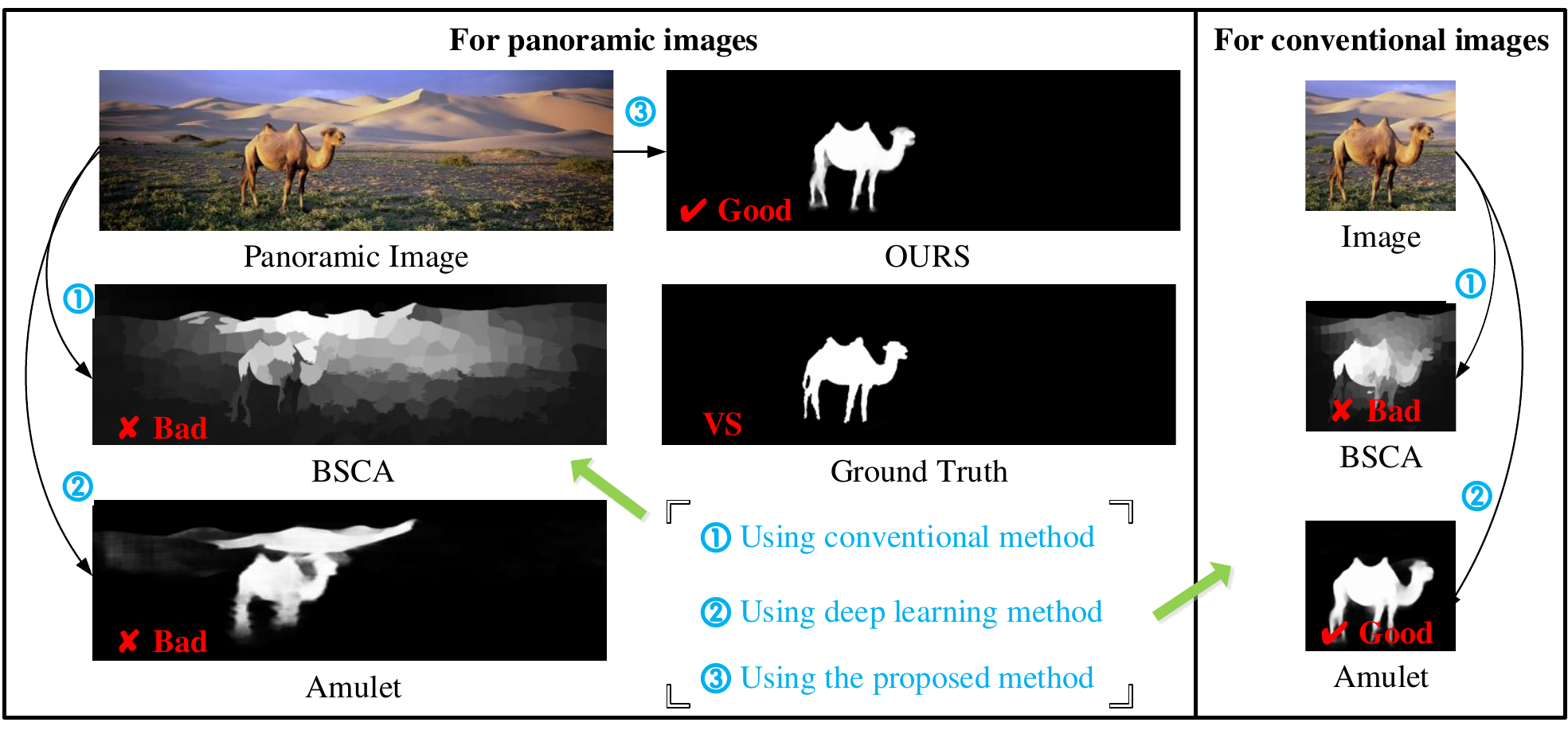}
\end{center}
\caption{The comparison of existing saliency detection algorithm between conventional images and panoramic images.}
\label{fig1}
\end{figure}

Early work~\cite{Itti1998A} on computing saliency aimed to model and predict human
gaze on images. Recently the field has expanded to include
the segmentation of entire salient regions or objects~\cite{Achanta2009}. Most works~\cite{8265566,10.1007/978-3-319-64698-5_2,8265388,Zhu2018,8265548,Zhu2018An,DBLP:journals/corr/abs-1711-00322} extract salient regions which exhibit highly
distinctive features compared to their surrounding regions, based on
the concept of center-surround contrast. Moreover, additional
prior knowledge for spatial layout of foreground objects and background
can be also used~\cite{Wei2012, Zhu2014, Margolin2013, Cheng2013}. These
assumptions have been successfully employed to improve the performance
of saliency detection for conventional images with common
aspect ratios. However, existing bottom-up methods~\cite{Qin2015Saliency} show somewhat poor performance for complex situations. With the development of neural networks, some algorithms~\cite{zhang2017amulet} adopt deep learning based methods to deal with this problem. This trend is analogous to the biological evolution process, which can be regarded as evolutionary theory of saliency detection (shown in the right of Fig.\ref{fig1}).

Recently, panoramic images, which yield wide fields of view,
become popular in all kinds of media contents and draw much attention
in many practical applications.
However, to accurately calculate visual saliency of panoramic images, both traditional methods and deep learning based methods proposed in recent years would fail in some cases of panoramic images, where complex background is present (shown in the left of Fig.\ref{fig1}). Thus we propose an effective saliency detection framework for panoramic images.

In this work, we propose an automatic salient object detection framework for panoramic images using a dual-stage region growing and fixation prediction model. Panoramic images exert several distinct characteristics compared to conventional
images.
We find that spatial density patterns is useful for images with high resolutions.
Therefore, we first employ a spatial density pattern detection method for the panoramic image to roughly extract the proposal regions, and use a dual-stage region growing process to get the proposed regions. Meanwhile, the eye fixation model is deployed in the framework to locate the focus of visual attention in images, which is inspired by the mechanisms of the human vision system.
Then, the previous saliency information is combined by the maxima normalization to get the coarse saliency map. Finally, a geodesic refinement is employed to derive the final saliency map.

The main contributions of this paper are listed as follows:

1. A new automatic salient object detection framework is proposed for panoramic images, which combines the usage of region growing and eye fixation model. Density map is first introduced in our work as a feature representation for saliency calculation.

2. To fairly evaluate the performance of the proposed method, we build a new high-quality panoramic dataset (SalPan) with an accurate ground truth annotation method which can eliminate the ambiguous of salient objects. This SalPan dataset will be released publicly after publication.

3. Compared with both conventional algorithms and deep learning based algorithms, the proposed method achieves the state-of-the-art performance on SalPan dataset and other large-scale salient object detection datasets including the recent ECSSD~\cite{Shi2016Hierarchical}, DUT-OMRON~\cite{Zhao2015Saliency} and SED~\cite{Borji2015What}. In addition, the proposed algorithm is fast on both modern CPUs and GPUs.

4. The proposed framework can also be tailored for small target detection.

To our thoughts, this research may help to exploit the perception characteristics of the human visual system
for large-scale visual contents over a wide field of view.

The rest of this paper is organized as follows. Section 2 briefly
reviews related work. Sections 3 explains the proposed framework. Section 4 presents the experimental results.
Section 5 discusses some practical guidelines. Section 6 concludes this paper.
\section{Related Works}
In this section, we have a brief review of classical eye fixation models, traditional saliency algorithms and deep learning based saliency detection methods.
\subsection{Eye Fixation Model}
The first models for saliency prediction were biologically
inspired and based on a bottom-up computational
model that extracted low-level visual features such as intensity,
color, orientation, texture and motion at multiple
scales. Itti et al.~\cite{Itti1998A} proposed a model that combines multiscale
low-level features to create a saliency map. Harel et
al.~\cite{Cerf2008} presented a graph-based alternative that starts from
low-level feature maps and creates Markov chains over various
image maps, treating the equilibrium distribution over
map locations as activation and saliency values.
Since the eye fixation model can mimic the  process of the human visual system, thus,
we embed the fixation prediction model into our framework.
\subsection{Traditional Saliency Algorithm}
Saliency detection for conventional images could be implemented
based on either top-down or bottom-up models.
Top-down models~\cite{Oliva2003,Gao2005,Gao2009,Yang2012,chen2017m} required high level interpretation usually provided by training sets in supervised learning. Contextual saliency was formulated
according to the study of visual cognition: global scene context of
an image was highly associated with a salient object~\cite{Oliva2003}. The most
distinct features were selected by information theory based methods~\cite{Gao2009}. Salient objects were detected by joint learning of a dictionary for object features and conditional random field classifiers
for object categorization~\cite{Yang2012}.
While these supervised approaches can effectively detect
salient regions and perform overall better than bottom-up
approaches, it is still expensive to perform the training
process, especially data collection.

In contrary, bottom-up models~\cite{Qin2015Saliency,Shi2016,ZhuICCV2017,Li2015,7966712,crm2016spl,crm2018tip} did not require
prior knowledge such as object categories, but obtained saliency
maps by using low level features based on the center-surround
contrast. They computed feature distinctness of a target region,
e.g., pixel, patch or superpixel, compared to its surrounding regions
locally or globally. For example, feature difference was computed
across multiple scales, where a fine scale feature map represented
the feature of each pixel while a coarse scale feature map described
the features of surrounding regions~\cite{Itti1998A}. Also, to compute center-surround
feature contrast, spatially neighboring pixels were assigned
different weights~\cite{Perazzi2012}, or random walk on a graph was used~\cite{Kim2014}.

Bottom-up based approaches did not need
data collection and training process, consequently requiring
little prior knowledge. These advantages make bottom-up
approaches more efficient and easy to implement in a wide
range of real computer vision applications.
A complete survey of these methods is beyond the scope of this paper and we refer the readers to a recent survey paper~\cite{Borji2014Salient} for details.
In this paper, we focus on the bottom-up approach.

\subsection{Deep Learning based Saliency Detection Method}
With the performance of deep convolutional neural achieving near human-level performance in image classification and recognition task. Many algorithms adopt deep learning based methods~\cite{zhang2017amulet,wang2017learning,Li2016Deep,He2016Delving,Wang2016Saliency,Li2015Visual,PDNet2018}. Instead of constructing hand-craft features, this kind of top-down methods have achieved state-of-the-art performance on many saliency detection datasets. However, deep learning based algorithms exist the following limitations: \textit{1) need a large number of annotated data for training. (2) very time-consuming in the learning process even with GPU of high computation ability. (3) training instances are not uniformly sampled. (4) sensitive to noise in training samples and image resolution.}

Bottom-up based approaches do not need
data collection and training process, consequently requiring
little prior knowledge. These make bottom-up approaches more suitable for real-time applications. Meanwhile, bottom-up methods are not sensitive to image scale, capacity and type. These advantages make bottom-up
approaches more efficient and easy to panoramic saliency detection. In this paper, we
focus on the bottom-up approach.

%
%
\begin{figure*}

\begin{center}
\includegraphics[width=6.636in,height =2in]{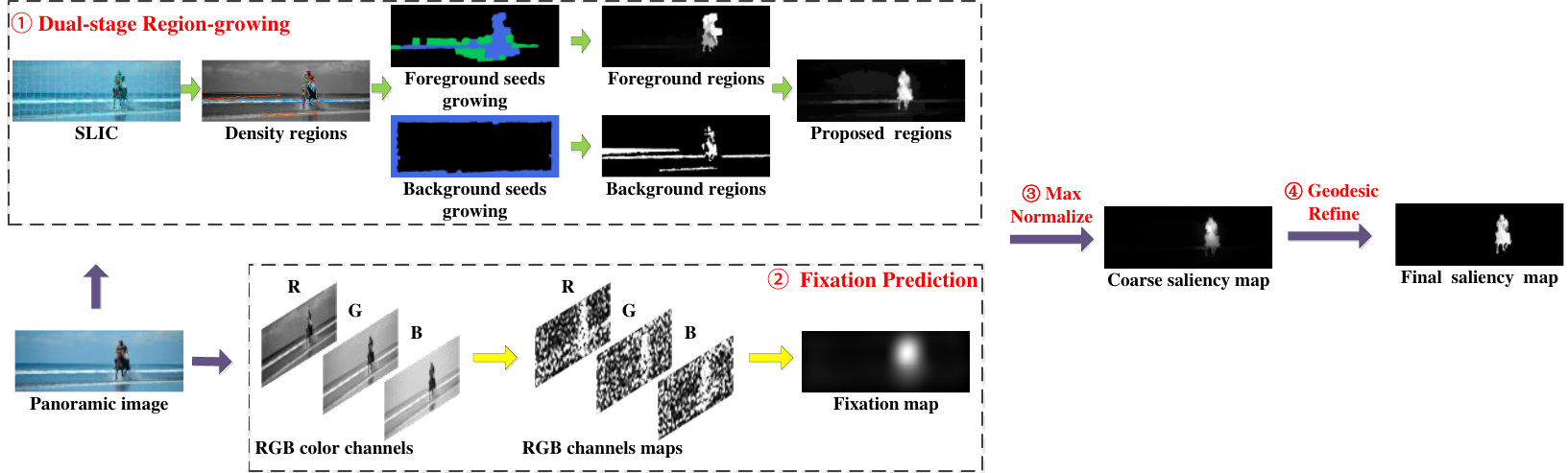}
\end{center}
\caption{The framework of the proposed algorithm. \textbf{Top:} Dual-stage region-growing method. \textbf{Bottom:}
Fixation prediction algorithm. The
temporal outputs of the two pathways are integrated using maxima normalization. The final output is refined by geodesic refinement. }
\label{fig2}
\end{figure*}
\section{The Proposed Framework}
Here we describe the proposed framework in four main steps. First, we employ a dual-stage region-growing algorithm for salient regions proposal. Second, an eye fixation prediction model is used to find visually attractive locations in the image, which is then combined with proposal regions in the previous step to form a reliable saliency map.
Third, a maxima normalization is utilized to optimize the previous saliency information.
Fourth, the final saliency map is obtained after a refinement step based on geodesic distance. The main framework is depicted in Fig. \ref{fig2}s.

\subsection{A Dual-stage Region-growing Based Detection}
In this section, we first use the efficient Simple Linear Iterative Clustering
(SLIC) algorithm~\cite{Lucchi2012Supervoxel} to segment the image into
smaller superpixel elements in order to capture the essential local structural
information of the image. Then , we roughly extract regions that have significant and different densities compared with its neighbors. After obtaining the proposal regions, we use a dual-stage seed estimation and growing to improve previous map. One is the foreground seeds based ranking process, the other is the background seeds based ranking process. Finally, we merge the two detection results to get the proposed salient regions.
\subsubsection{Density Map}
The notion of density comes from fractal theory. Density
measures locally the variation of a quantity over a number
of scales. It has been used before for texture classification~\cite{Miller2009Viewpoint}. The idea is that at small scales naturally
occurring objects (here the texture) change over scale in a
way, that can be modeled as an exponential function of the
scale. Thus the exponent of the function is a good statistical
descriptor of the object which is invariant to a wide
range of spatial transformations. In this work we focus on
the density variation property of the density map.

Let $I(x)$ be a grayscale image and let $\mu(x, r)$ be a measure
on $R^2$. For our purposes we choose $\mu(x, r)$ to be the
sum of image intensities in a disk of radius $r$ around point
$x$, i.e. $\mu(x, r) = \sum_{||y-x||\leq r}I(y)$. We use the power law to
express $\mu$ as a function of $r$:
 \begin{equation}
\mu(x, r) = kr^{d(x)} ,
\end{equation}
 \begin{equation}
\log (\mu(x, r)) = \log k + d(x)\log r ,
\end{equation}
 \begin{equation}
d(x) = \liminf \frac{\log (\mu (x,r))}{\log r} .
\end{equation}

We define the exponent $d(x)$, also known as Holder exponent, to be the local density function of image $I(x)$ at point
$x$. Intuitively, it measures the degree of regularity of intensity
variation in a local neighbourhood around point $x$.

The density map
values are the same within regions of different intensity as
well as within regions of smoothly varying intensity. In essence, the density map preserves important textural
features by responding to abrupt intensity discontinuities
and avoiding smoothly varying regions. In our work, the obtained density map is utilized as a feature map for coarse salient region proposals. We adopt the region-based contrast method for this purpose. GrabCut~\cite{Rother2004}is used to segment the image into a number of regions, then saliency computation is performed on these regions. The saliency value of each regions is derived as follows:
\begin{equation}
S_b(r_k) = \sum_{r_k\neq r_i}w(r_i)D_r(r_k,r_i) ,
\end{equation}
where $w(r_i)$ is the weight of region $r_i$ and $D_r(\cdot)$ is the density distance metric between the two regions.
\begin{equation}
D_r(r_1,r_2) = \sum_{i=1}^{n_1} \sum_{j=1}^{n_2}f(d_1,i)f(d_2,j)D(d_1,i,d_2,j) ,
\end{equation}
where $f(d_k,i)$ is the probability of the $i$-th density $d_{k,i}$ among all $n_k$ densities in the $k$-th region $r_k$, $k = \{1,2\}$.

In the end, the map is binarized to obtain a mask, which could be treated as object region proposals. The mask map is denoted $S_b$ .
\subsubsection{Foreground Seeds based Region Growing}
We denote the map generated by previous section, $S_{b}$, which indicates proposed density regions.

Unlike previous works\cite{Li2013Saliency,Yang2013Saliency} that treat some regions as certain seeds, we provide a more flexible scheme for seeds estimation. We define two types of seed elements: strong seeds and weak seeds. Strong seeds have high probability of belonging to foreground/background while weak seeds have relatively low probability of belonging to foreground/background. For foreground seeds, the two types of seeds are selected by:
\begin{equation}
\label{fore1}
C_{fore}^+ = \{  i | S_{p}(i) >= 2 \cdot mean(S_{p}) \} ,
\end{equation}
\begin{equation}
\label{fore2}
\begin{aligned}
C_{fore}^- = \{  i | &S_{p}(i) >= mean(S_{p}) ~~and~~ \\
&S_{p}(i) < 2 \cdot mean(S_{p}) \} ,
\end{aligned}
\end{equation}
where $C^+$ denotes the set of strong seeds and $C^-$ weak seeds,
$i$ represent $i$th superpixel. mean(.) is the averaging function. It is obvious from formula (\ref{fore1})(\ref{fore2}) that elements of higher degree of surroundedness are more likely to be chosen as strong foreground seeds, which is consistent with intuition.

For saliency calculation based on given seeds, a ranking method in ~\cite{Zhou2013Ranking} that exploits the intrinsic manifold structure of data for graph labelling is utilized. The ranking method is to rank the relevance of every element to the given set of seeds.
We construct a graph that can represent an whole image as in work~\cite{Yang2013Saliency}, where each node is a superpixel generated by SLIC.
%

The ranking procedure is as follows: Given a graph $G = (V, E)$ ,where the nodes are $V$ and the edges $E$ are weighted by an affinity matrix $\boldsymbol{W} = [w_{ij}]_{n \times n}$. The degree matrix is defined by $\boldsymbol{D} = diag\{  d_{11}, ... , d_{nn} \}$, where $d_{ii} = \sum_j w_{ij}$.
The ranking function is given by:
\begin{equation}
\boldsymbol{g}^* = ( \boldsymbol{D} - {\alpha}{\boldsymbol{W}} )^{-1} \boldsymbol{y}  \label{eq} .
\end{equation}
The $\boldsymbol{g}^*$ is the resulting vector which stores the ranking results of each element. The $\boldsymbol{y} = [y_1, y_2, ... ,y_n]^{T}$ is a vector indicating the seed queries.

 In this work, the weight between two nodes is defined by:
\begin{equation}
\label{fore4}
w_{ij} = e^{-\frac{\| c_i - c_j \|}{ {\sigma}^2 }} ,
\end{equation}
where $c_i$ and $c_j$ denote the mean of the superpixels corresponding to two nodes in the CIE LAB color space, and $\sigma$ is a constant that controls the strength of the weight.

Different from \cite{Yang2013Saliency} that define $y_i = 1$ if $i$ is a query and $y_i = 0$ otherwise, we define $y_i$ as the strength of the query extra. That is, $y_i = 1$ if $i$ is a strong query, and $y_i = 0.5$ if $i$ is a weak query, and $y_i = 0$ otherwise.

For foreground seeds based ranking, all elements are ranked by formula (\ref{fore4}) given the sets of seeds in (\ref{fore1})(\ref{fore2}).
\subsubsection{Background Seeds based Region Growing}
Complementary to foreground seeds estimation and growing, background seeds estimation and growing aims to extract regions that are different from background in feature distribution.
We first select a set of background seeds and then calculate saliency of every image element according to its relevance to these seeds. We divide the elements on image border into two categories(strong seeds and weak seeds) as in foreground situation.
%
We denote the average value of all border elements as $\overline{c}$.
The euclidean distance between each feature vector and the average feature vector is computed by ${\bf{d}}_c = \bf{dist} (c, \overline{c})$,
 the average of ${\bf{d}}_c $ is denoted by $\overline{{\bf{d}}_c}$. The background seeds are estimated by:

\begin{equation}
C_{back}^+ = \{  i | {\bf{d}}_c (i)  >= 2 \cdot\overline{{\bf{d}}_c} \} ,
\end{equation}
\begin{equation}
C_{back}^- = \{  i | {\bf{d}}_c (i)  >= \overline{{\bf{d}}_c} ~~ and~~  {\bf{d}}_c (i)  < 2 \cdot \overline{{\bf{d}}_c}  \} ,
\end{equation}
where $C_{back}^+$ denotes strong background seeds, $C_{back}^-$ denotes weak background seeds.

Similar to foreground situation, the value of indication vector for background seeds $\boldsymbol{y}$ is $y_i = 1$ if $i$ belongs to $C_{back}^+$,
$y_i = 0.5$ if $i$ belongs to $C_{back}^-$
and 0 otherwise.
Relevance of each element to background seeds is computed by formula (\ref{eq}).
Elements in resulting vector ${\boldsymbol{g}}^*$ indicates the relevance of a node to the background queries, and its complement is the saliency measure.The saliency map using these background seeds can be written as:
\begin{equation}
S(i) = 1 - {\boldsymbol{g}}^*(i)   \qquad i=1,2,...,N.
\end{equation}

The background saliency map and foreground saliency map is combined by multiplication to form a coarse saliency map, as shown in the end of stage 1 in Fig.\ref{fig2}.

\subsection{Fixation prediction}
Whether a location is salient or not largely depends on how much it attracts human attention. A large number of recent works on eye fixation prediction have revealed more or less the nature of this issue. Eye fixation prediction models simulate the mechanisms of the human visual system, and thus can predict the probability of a location to attract human attention. So in this section, we use eye fixation model to help us ensure which region has more power to grab human attention.

Panoramic images are often with wide fields of view, and consequently are computationally more expensive compared with conventional images. Algorithms based on color contrast, local information are not suitable for being a preprocessing step for panoramic images, for these algorithms are time-consuming and would spend a lot of computational resources. Thus we are seeking a more efficient method to help us to rapidly scan the image and roughly locate where attract human attention. Obviously Fixation prediction models in frequency domain fit this demand, for these models are computationally efficient and easy to implement.

The signature model approximately isolate the spatial support of foreground by taking the sign of the mixture signal x in the transformed domain and then transform it back to spatial domain, i.e., by computing the reconstructed image $\overline{\textbf{x}} = IDCT[sign(\widehat{\textbf{x}})]$. $\widehat{\textbf{x}}$ stands for DCT transform of $x$. The image signature is defined as
\begin{equation}
IS(\textbf{x}) = sign(DCT(\textbf{x})) .
\end{equation}
And the saliency map is formed by smoothing the squared reconstructed image defined above
\begin{equation}
S_m = g * (\overline{\textbf{x}}\circ \overline{\textbf{x}}) ,
\end{equation}
where $g$ is a Gaussian kernel.

The image signature is a simple yet powerful descriptor of natural scenes, and it can be used to approximate the spatial location of a sparse foreground hidden in a spectrally sparse background. Compared with other eye fixation models, image signature has a more efficient implementation, which runs faster than all other competitors.

To combine the proposed salient regions in previous section with saliency map $S_m$ produced by image signature, we assign the saliency value of the proposed salient regions by averaging the saliency values of all its pixels inside. For convenience we denote the resulted saliency map as $S_p$. That is, for a proposed region $p$, its saliency value is defined as
\begin{equation}
S_p(i) = ( \sum_{i\in p} S_m(i) ) / A(p)   \quad ,  i \in p ,
\end{equation}
where $A(p)$ denote the number of pixels in the $p$th region.

\begin{figure}[!h]

\begin{center}
\includegraphics[width=3.2in,height =2.06in]{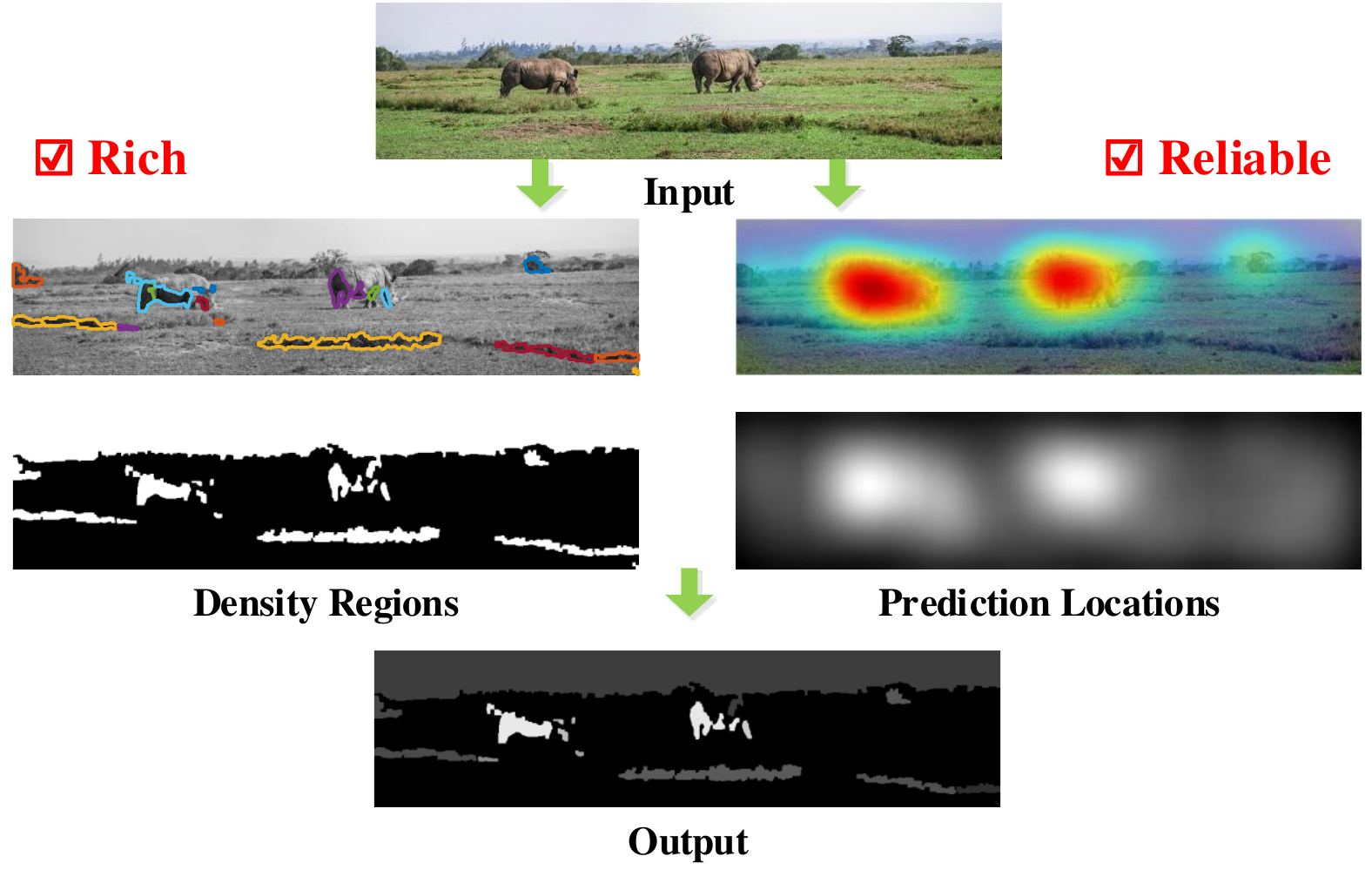}
\end{center}
\caption{The illustration of complementary between density detection and fixation prediction.}
\label{fig3}
\end{figure}

As shown in Fig.\ref{fig3}, fixation prediction model reliably locate  visually attractive positions, while density detection pops out a number of proposal regions. One is for reliable but fuzzy locating saliency, while the other is for rich pop-out object proposals. These two step are therefore complementary.
\subsection{Maxima Normalization}
Fusing saliency detection results of multiple models has
been recognized as a challenging task since the candidate
models are usually developed based on different cues or assumptions.
Fortunately, in our case, the integration problem
is relatively easier since we only consider the outputs from
two pathways. Since there is no prior knowledge or other
top-down guidance can be used, it is safer to utilize the map
statistics to determine the importance of each pathway. Intuitively
in the final integration stage, we combine the results
from two pathways by summing them after Maxima Normalization
(MN) (Algorithm 1).
\begin{algorithm}[!htb] %

\caption{ Maxima Normalization $N_{max}(S_p, t)$} %
\hspace*{0.02in}{\bf Input:} Previous map $S_p$, thresh of local maxima $t = 0.1$;

\hspace*{0.02in}{\bf Output:} Normalized Saliency Map $S_N$;

\begin{algorithmic}[1] %
\STATE Set the number of maxima $N_M = 0$ ;
\STATE Set the sum of the maxima $V_M = 0$ ;
\STATE Set Global Maxima $G_M = max(S)$ ;
\STATE \textbf{for} all pixel (x, y) of S \textbf{do} :
\STATE \quad\textbf{if} $S(x, y) > t$ \textbf{then}
\STATE \quad$R = {S(i, j)| i = x - 1, x + 1, j = y - 1, y + 1}$ .
\STATE \qquad\textbf{if} $S(x, y) > max(R)$ \textbf{then}
\STATE \qquad$V_M = V_M + S(x, y)$ .
\STATE \qquad$N_M = N_M + 1$ .
\STATE \qquad{\bf end for}
\STATE \quad{\bf end for}
\STATE {\bf end for}
\STATE $S_N = S\cdot(G_M - V_M / N_M)^2 / G_M$ .
\RETURN \textbf{return} Normalized map $S_N$
\label{alg1}
\end{algorithmic}
\end{algorithm}

The Maxima Normalization operator $N_{max}$(.) was originally
proposed for the integration of conspicuous maps from
multiple feature channels~\cite{Itti1998A},
which has been demonstrated very effective and has a very
convincing psychological explanation.
\subsection{Geodesic refinement}
The final step of our proposed approach is refinement with geodesic distance ~\cite{Tenenbaum2000}. The motivation underlying this operation is based on thought that determining saliency of an element as weighted sum of saliency of its surrounding elements, where weights are corresponding to Euclidean distance, has a limited performance in uniformly highlighting salient object. We tend to find a solution that could enhance regions of salient object more uniformly. From recent works~\cite{Zhu2014}, we found the weights may be sensitive to geodesic distance.

The input image is first segmented into a number of superpixels based on linear spectral clustering method~\cite{Perazzi2012} and the posterior probability of each superpixel is calculated by maxima normalization operation $S_N$ of all its pixels inside. For $j$th superpixel, if its posterior probability is labeled as
$S_N(j)$, thus the saliency value of the $q$th superpixel is refined by geodesic distance as follows:
\begin{equation}
S(q) = \sum_{j=1}^{J} w_{qj}\cdot S_N(j) ,
\end{equation}
where $J$ is the total number of superpixels, and $w_{qj}$ will be a weight based on geodesic distance~\cite{Zhu2014} between $q$th superpixel and $j$th superpixel.

First, an undirected weight graph has been constructed connecting all adjacent superpixels $(a_{k},a_{k+1})$ and assigning their weight $d_c(a_{k}, a_{k+1})$ as the Euclidean distance between their saliency values which are derived in the previous section. Then the geodesic distance between two superpixels $d_g(p, i)$ can be defined as accumulated edge weights along their shortest path on the graph:
\begin{equation}
d_g(p,i) = \min_{a_1 = p, a_2, a_3,..., a_n = i} \sum_{k=1}^{n-1} d_c(a_k, a_{k+1}) .
\end{equation}
In this way we can get geodesic distance between any two superpixels in the image. Then the weight $\delta_{pi}$ is defined as
\begin{equation}
w_{qj} = exp\lbrace - \frac{d_{g}^{2}(p,i)}{2\sigma_c^{2}} \rbrace ,
\end{equation}
where $\sigma_c$ is the deviation for all $d_c$ values. From formula (6) we can easily conclude that when  $p$ and $i$ are in flat region, saliency value of $i$ would have a higher contribution to saliency value of $p$, and when $p$ and $i$ are in different regions between which a steep slope is existed, saliency value of $i$ tends to have a less contribution to saliency value of $p$. 
%

Since an object often contains some
homogenous parts, the initial saliency value of a superpixel
could be spread to the other connected homogenous parts, indicating
the propagation may be achieved through connectivity.
This can also be observed in the background where image
background can be divided into large homogenous regions.
Noting that the coarse saliency could render more saliency to
the target object while less to the background (Fig.2). Hence,
low saliency in background part could be spread over the
entire background region after propagation. Eventually, the
background could be suppressed to very low saliency values.

\section{Experiments}
To fairly evaluate the performance of the proposed framework, we build a new dataset of panoramic landscape
images (SalPan), and evaluate the performance of the proposed
saliency detection algorithm compared with 12 state-of-the-arts methods. More experimental analysis on the effectiveness of our method are given as follows.
\subsection{Datasets}
We collect a new panoramic dataset SalPan composed of 123
panoramic images.
In general, saliency may be ambiguous in
images with highly complex scenes and wide fields of view,
thus we propose an accurate ground truth annotation method to eliminate the ambiguous.

To build a ground-truth for
SalPan dataset we first conducted a comprehensive user study.
To identify where participants were looking while watching
the films, we monitor their eye movements using a Gazepoint
GP3 Eye Tracker. Image presentation was controlled
using the Gazepoint Analysis Standard software.
For the purpose of the study we recruited 10 participants
(5 males, 5 females). Ages ranged from 20 to 57 with
the mean age of 25. All the participants had normal or
corrected-to-normal vision and were naive to the underlying
purposes of the experiment. Then, we record the tracking path and allocate values to each eye movement location according to the times of visual attention.
Finally, we pick up those values above the average and label the ground truth with complete objects corresponding attention points.

Meanwhile, we make the image acquisition and image annotation independent to each other, we can avoid dataset design bias, namely a specific type of bias that is caused by experimenters' unnatural selection of dataset images.
After picking up the salient regions, we adhere to the following rules to build the ground truth:

$\bullet$ disconnected regions of the same objects are labeled separately;

$\bullet$ solid regions are used to approximate hollow objects, such as bike wheels.

\begin{figure}
\begin{center}
\includegraphics[width=3.2in,height =1.345in]{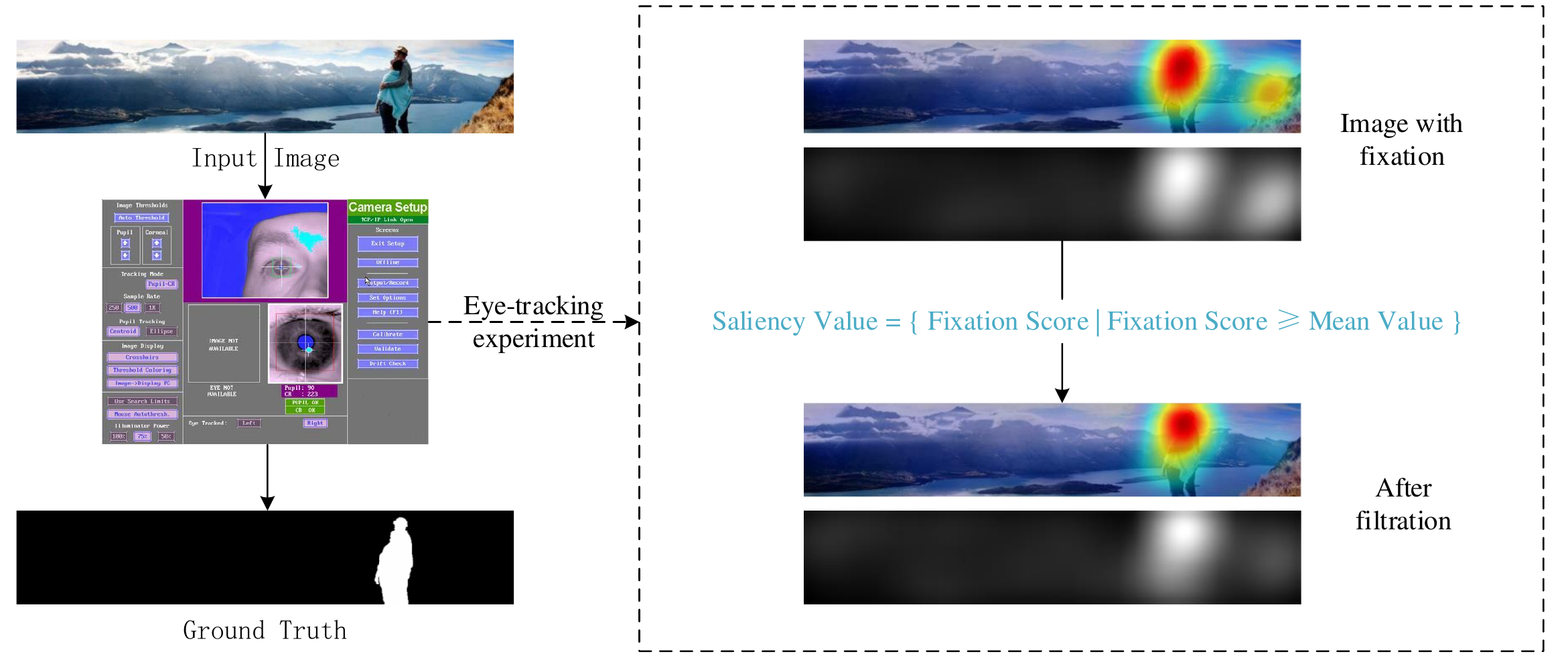}
\end{center}
\caption{The process of generating the ground truth for the SalPan dataset.}
\label{fig4}
\end{figure}

The process of labeling the ground truth is shown in Fig. \ref{fig4}.

\subsection{Evaluation indicators}
Experimental evaluations are based on standard measurements including precision-recall curve, precision value, recall value, AUC value, MAE (Mean Absolute Error) value, time-consuming with coding type and F-measure.

Besides, We also adopt a new
structural similarity measure known as Structure-measure (S-measure, proposed in ICCV17)~\cite{fan2017structure}
which simultaneously evaluates region-aware and object aware
structural similarities between a saliency map and a
ground-truth map.

The precision is defined as:
\begin{equation}
Precision=\frac{\|{p_i\mid d(p_i)\geq d_t}\cap{p_g}\|}{\|{p_i\mid d(p_i)\geq d_t}\|} ,
\end{equation}
where ${p_i\mid d(p_i)\geq d_t}$ indicates the set that binarized from a saliency map using threshold $d_t$. ${p_g}$ is the set of pixels belonging to groundtruth salient object.

The recall is defined as:
\begin{equation}
Recall = \frac{\|{p_i\mid d(p_i)\geq d_t}\cap{p_g}\|}{\|{p_g}\|} .
\end{equation}
The precision-recall curve is plotted by connecting the P-R scores for all thresholds.

The MAE is formulated as:
\begin{equation}
MAE=\frac{\sum_{i=1}^N \|GT_i-S_i\|}{N} ,
\end{equation}
where $N$ is the number of the testing images, $GT_i$ is the area of the ground truth of an image $i$, $S_i$ is the area of the result of an image $i$.

The F-measure is formulated as:
\begin{equation}
F-measure=\frac{(1+\beta^2)\times Precision\times Recall}{\beta^2\times Precision+Recall} ,
\end{equation}
where $\beta^2$ is set to 0.3 as did in many literatures.

The S-measure is formulated as:
\begin{equation}
S-measure=S = \alpha \times S_o + (1 - \alpha) \times S_r ,
\end{equation}
where $S_o$ and $S_r$ are region-aware and object-aware structural similarity
evaluation value, respectively.

\begin{figure*}
\begin{center}
\includegraphics[width=7in,height =1.5in]{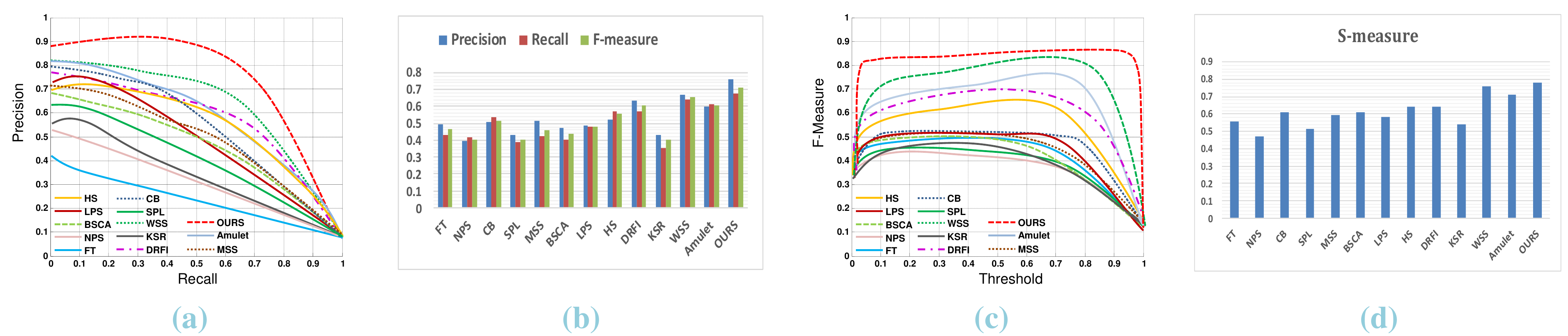}
\end{center}
\caption{Quantitative evaluation on SalPan dataset. (a)shows the PR curve. (b)shows the comparison of precision, recall and F-measure scores. (c)shows F-measure curve. (d)shows S-measure scores.}
\label{pr}
\end{figure*}
\begin{figure*}
\label{fig5}
\begin{center}
\includegraphics[width=7in,height =2.381in]{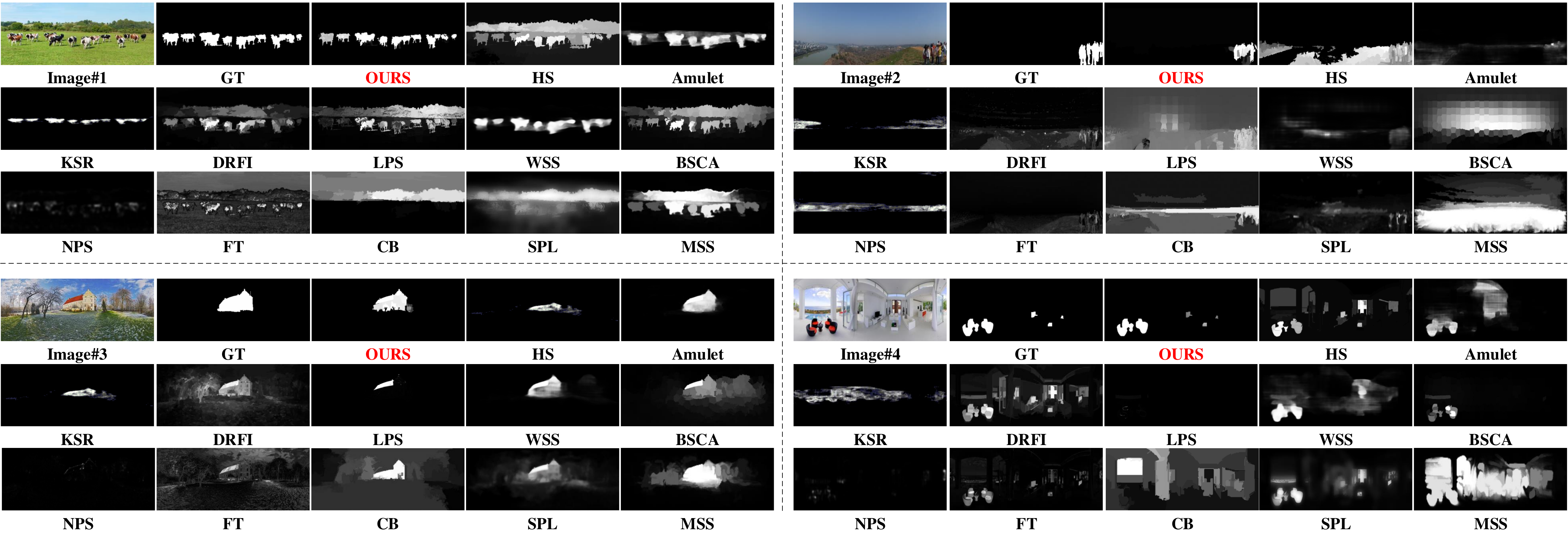}
\end{center}
\caption{Visual comparisons of different saliency algorithms on SalPan dataset.}
\label{visualpan}
\end{figure*}

\subsection{Ablation study}
We first validate the effectiveness of each step in our method: foreground seeds based region-growing (R-F), background seeds based region-growing (R-B),a dual-stage region-growing based detection (R-M), fixation prediction saliency detection (FP), maxima normalization fusion (MN) and geodesic refinement (GR). Table.I shows the validation results on SalPan dataset. We can clear see the accumulated processing gains after each step, and the final saliency results shows a good performance. After all, it proves that each steps in our algorithm is effective for generating the final saliency maps.
\begin{table}[!h]

\begin{center}
\begin{tabular}{|c|c|c|c|c|c|c|c|}
\hline
\multirow{2}{*}{*}&\multicolumn{3}{|c|}{Step 1}&\multicolumn{1}{|c|}{Step 2}&\multicolumn{1}{|c|}{Step 3}&\multicolumn{1}{|c|}{Step 4}\\
\cline{2-7}
 & R-F & R-B & R-M & FP & MN & GR\\
\hline
MAE &.291&.287&.271&.311&.256&.231\\
F-m &.689&.694&.715&.646&.734&.767\\
S-m &.727&.731&.737&.694&.752&.783\\
\hline
\end{tabular}
\label{tb3}
\end{center}
\caption{The results of MAE, F-measure and S-measure at each steps of the proposed algorithm. Step 1 and Step 2 are the concurrent procedure.}
\end{table}

Fig. \ref{visual6} show two group visual process of each steps of the proposed method. Step 1: Dual-stage Region growing. Step 2: Fixation Prediction. Step 3: Max Normalization. Step 4: Geodesic Refinement. From the process shown in the figure, we can see the details of our framework and how each step contribute to the final result.

\begin{table*}
\label{tb1}
\begin{center}
\begin{tabular}{|c|c|c|c|c|c|c|c|c|c|}
\hline
* & MAE Value& AUC Value& Precision Value& Recall Value& F-measure& S-measure& Time(s) & Code Type\\
\hline
FT & 0.3868 & 0.5072 & 0.4942 & 0.4286&0.4643& 0.5580 & \color{red}\textbf{0.593} & C\\
\hline
NPS & 0.3613 & 0.4213 &0.3973  & 0.4195 &0.4053&0.4722& 2.397 & M $\&$ C \\
\hline
CB & 0.4238 & 0.6515 & 0.5112 & 0.5398 &0.5125&0.6095& 4.196 & M $\&$ C \\
\hline
SPL & 0.3125 & 0.4517 & 0.4324 & 0.3861 &0.4060&0.5156& 1.926 & M $\&$ C \\
\hline
MSS & 0.4627 & 0.5221 & 0.5185 & 0.4234 &0.4621&0.5912& 3.893   & M $\&$ C \\
\hline
BSCA & 0.4137 & 0.5115 & 0.4716 & 0.4067 &0.4360&0.6079& 1.876 & M $\&$ C\\
\hline
LPS & 0.3697 & 0.5184 & 0.4867 & 0.4794 &0.4810&0.5812& 2.193  &M $\&$ C\\
\hline
HS & 0.4090 & 0.6816 & 0.5212 & 0.5692 &0.5597&0.6395& 2.896 & M $\&$ C \\
\hline
DRFI & 0.3592  & \color{blue}\textbf{0.7021} & 0.6319 & 0.5710 &0.6072 &0.6402& 1.903  &M $\&$ C\\
\hline
KSR & 0.3222 & 0.4608 & 0.4345 & 0.3507 & 0.4056 &0.5397&2.537& M $\&$ C\\
\hline
WSS& \color{blue}\textbf{0.3090} & 0.6507 & \color{blue}\textbf{0.6687} & \color{blue}\textbf{0.6380} & \color{blue}\textbf{0.6546} &\color{blue}\textbf{0.7579}&2.751& M $\&$ C\\
\hline
Amulet & 0.3179 & 0.6321 &0.5962  & 0.6119 & 0.6069 &0.7121&3.216& M $\&$ C \\
\hline
OURS &  {\color{red}\textbf{0.2744}} &{\color{red}\textbf{0.8024}} & {\color{red}\textbf{0.7614}} & {\color{red}\textbf{0.6753}}&{\color{red}\textbf{0.7142}}&{\color{red}\textbf{0.7832}}& \color{blue}\textbf{1.075} & M $\&$ C \\
\hline
\end{tabular}
\end{center}
\caption{Evaluation indicators on SalPan dataset. M represents Matlab, C represents C++. The best two scores are shown in {\color{red}\textbf{red}} and {\color{blue}\textbf{blue}} colors, respectively.}
\end{table*}

\subsection{Comparison}
To illustrate the effectiveness of our algorithm, we compare our proposed methods with other 12 state-of-the-art ones including 9 conventional algorithms (FT~\cite{Achanta2009}, NPS~\cite{Hou2009}, CB~\cite{Jiang2011Automatic}, SPL~\cite{Liu2014Superpixel}, MSS~\cite{Tong2014Saliency}, BSCA~\cite{Qin2015Saliency}, LPS~\cite{Li2015}, HS~\cite{Shi2016}, DRFI~\cite{Wang2017}) and 3 deep learning based algorithms (KSR~\cite{wang2016kernelized}, WSS~\cite{wang2017learning} and Amulet~\cite{zhang2017amulet}). We use the codes provided by the authors to reproduce their experiments. For all the compared methods, we use the default settings suggested by the authors.

Fig. \ref{pr} compares the PR curves, where we see that
the proposed algorithm achieves a much higher performance than
that of the existing methods.

As observed in Fig. \ref{pr}, we see that the proposed algorithm has a higher F-measure score and S-measure score than any other competitors.

We measure the MAE value, precision value, recall value, AUC value, F-measure value and S-measure value using a resulting saliency map against the
ground truth saliency map, which are shown in Table.II. We have seen that
the proposed algorithm also achieves the best performance and
outperforms all other compared methods.

We also compare the average execution times of the proposed
algorithm and the other methods in Table II. Most of the methods
including the proposed one are implemented using MATLAB and
executed on an Intel i7 3.4 GHz CPU with 16 GB RAM. Results show that
most of the existing methods consume more time than the proposed algorithm.

In summary, from the comparison, we can conclude that our saliency results are more robustness and efficient on SalPan dataset.
Besides, the visual comparisons shown in Fig. \ref{visualpan} clearly demonstrate the advantages of the proposed method. We can see that our method can extract both single and multiple salient objects precisely. In contrast, the compared methods may fail in some situations.


\section{Discussion}
\textbf{Comparison in conventional datasets.} To further fairly verify the performance of the proposed method. We also compare our algorithm with other 12 state-of-the-art ones on 3 public saliency detection datasets, including ECSSD~\cite{Shi2016Hierarchical}, DUT-OMRON~\cite{Zhao2015Saliency} and SED~\cite{Borji2015What}.

Visual comparisons of salient region detection results are shown in Fig. \ref{visual5}. GT represents Ground Truth. The proposed method also shows good results on conventional datasets, which is in consistent with panoramic datasets.
\begin{figure}[!h]
\begin{center}
\includegraphics[width=3.4in,height =1.62in]{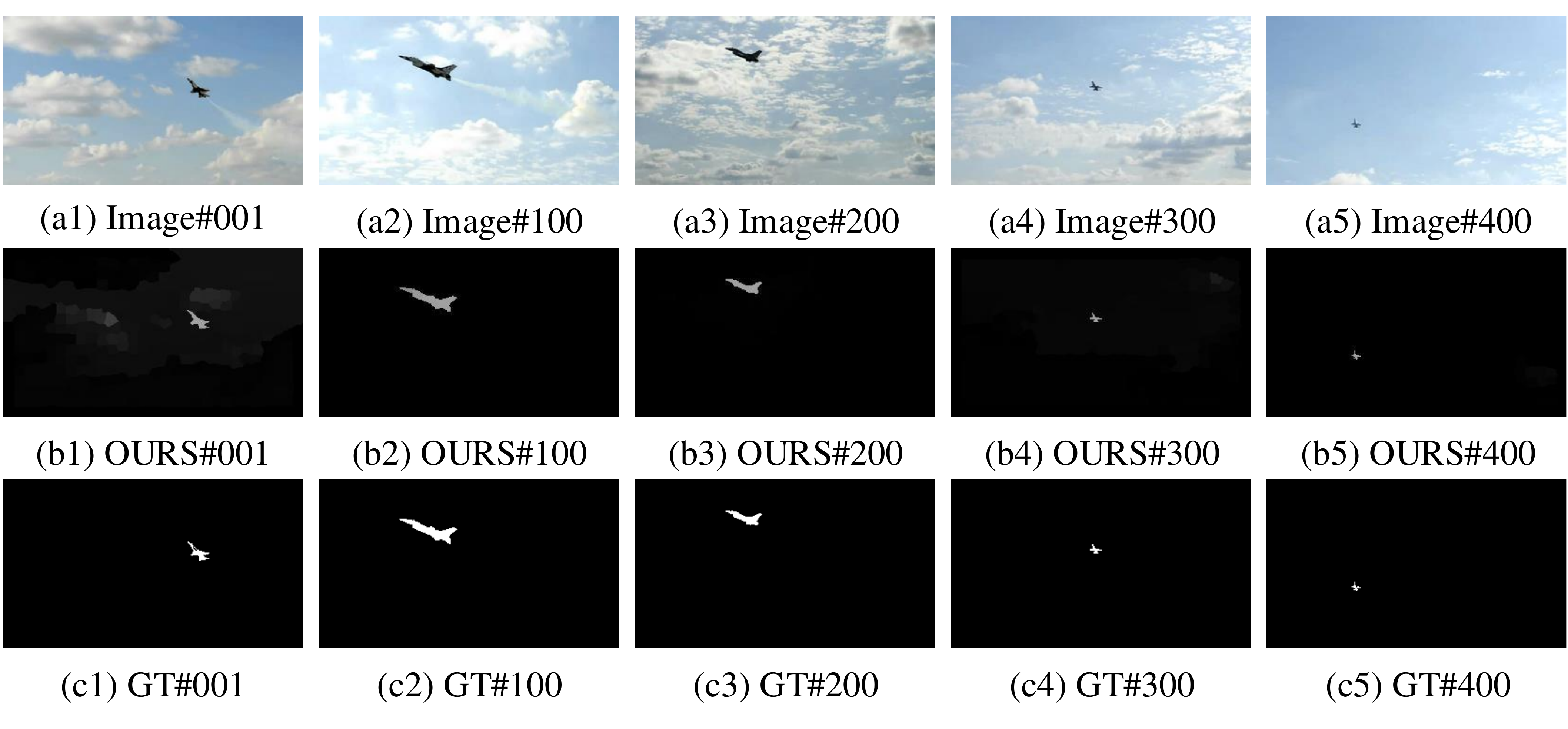}
\end{center}
\caption{The proposed algorithm is applied in small target detection. (a1)-(a5) represent different frames of original video.(b1)-(b5)
represent different frames of the proposed algorithm detection
results. (c1)-(c5) represent different frames of the ground truth.}
\label{fig:short}
\end{figure}

\textbf{Small target detection.} It is very interesting to find that the proposed automatic salient object detection algorithm for panoramic images using region growing and eye fixation model
is also valid for small target detection. We present a part
of our experimental results on the small target dataset~\cite{Lou2016}.
Small target detection plays an important role in many computer
vision tasks, including early alarming system, remote
sensing and visual tracking. The experimental results by
applying the proposed algorithm to small target detections in Fig. 5, respectively, which support our claim.

The underlying reason why the proposed algorithm can be applied
in small target detection and conventional images' saliency detection is that: the panoramic images also contain small objects and big objects, so, the image with the small target and conventional image can be seen as the part of its.

Therefore, we claim that the proposed automatic salient object detection algorithm is not only confined to panoramic images but
also conventional images and the images with small target.

\begin{figure*}
\begin{center}
\includegraphics[width=7in,height =6.2333in]{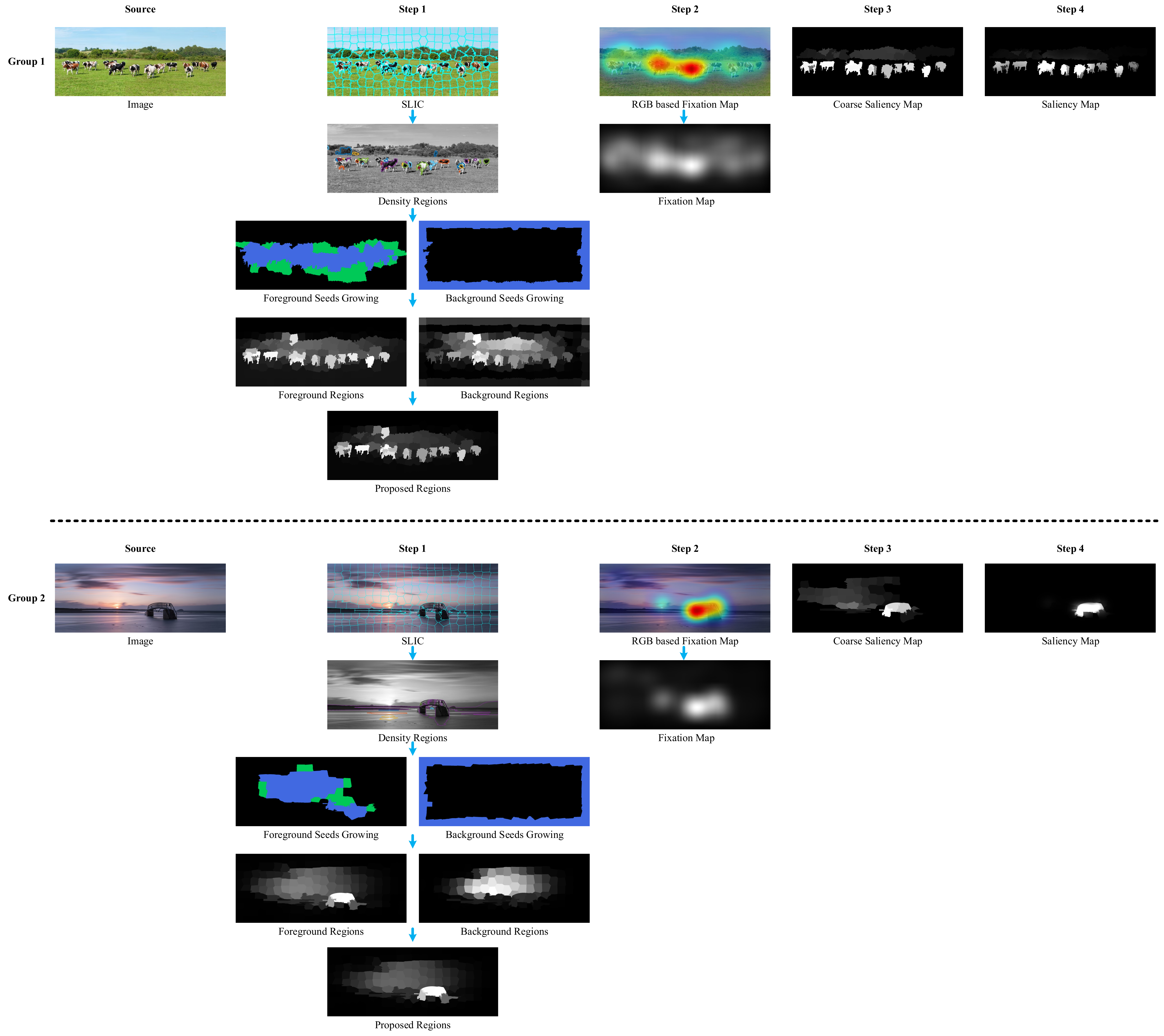}
\end{center}
\caption{Visual process of detailed steps of the proposed method. Step 1: Dual-stage Region growing. Step 2: Fixation Prediction. Step 3: Max Normalization. Step 4: Geodesic Refinement.}
\label{visual6}
\end{figure*}
\begin{figure*}
\begin{center}
\includegraphics[width=6.5in,height =8.6in]{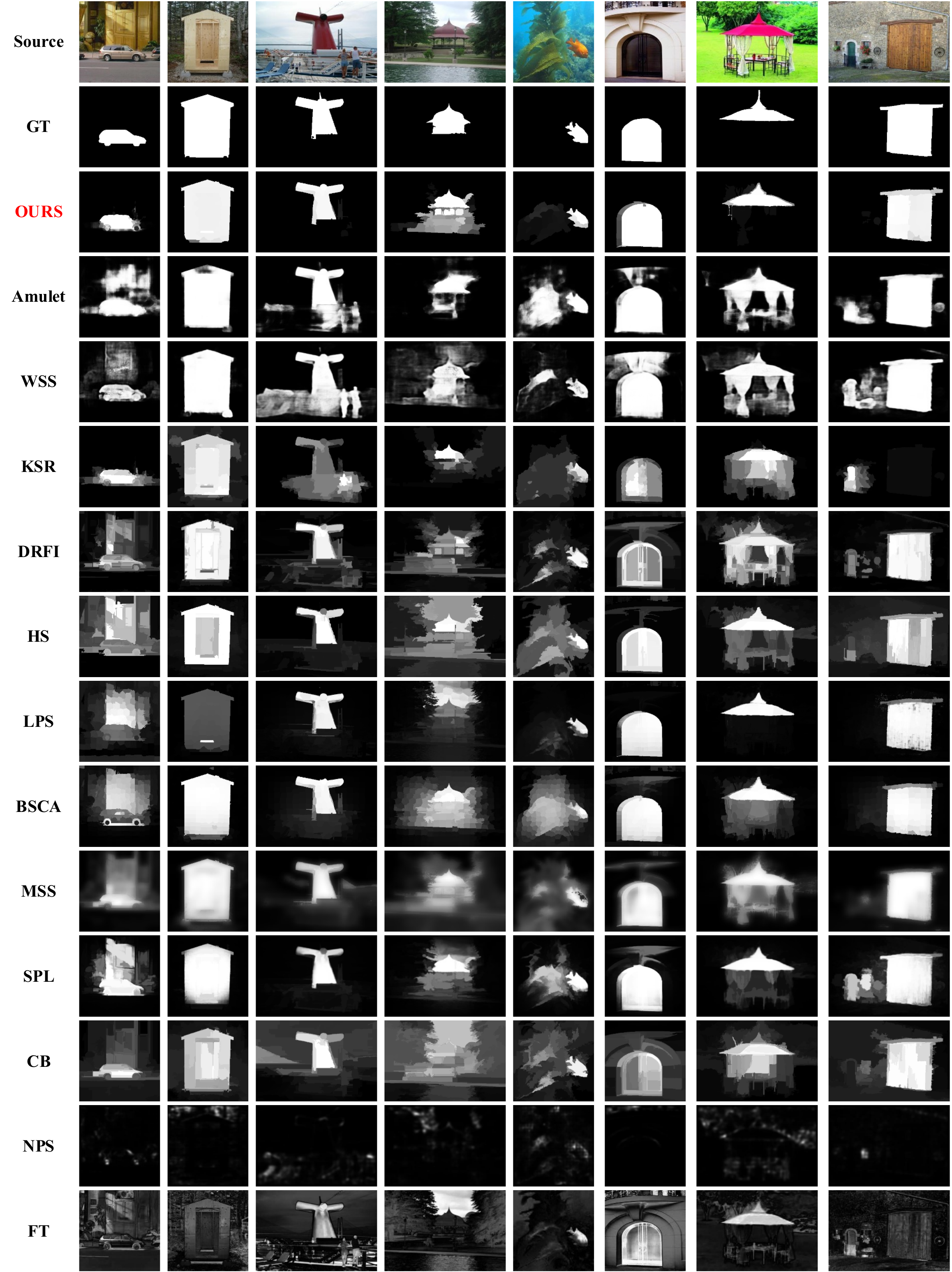}
\end{center}
\caption{Visual comparison with 12 state-of-the-art ones including 3 deep learning based algorithms (KSR~\cite{wang2016kernelized}, WSS~\cite{wang2017learning}, Amulet~\cite{zhang2017amulet}) and 9 conventional algorithms (FT~\cite{Achanta2009}, NPS~\cite{Hou2009}, CB~\cite{Jiang2011Automatic}, SPL~\cite{Liu2014Superpixel}, MSS~\cite{Tong2014Saliency}, BSCA~\cite{Qin2015Saliency}, LPS~\cite{Li2015}, HS~\cite{Shi2016}, DRFI~\cite{Wang2017}). }
\label{visual5}
\end{figure*}
\textbf{Limitation of the proposed method.} Since our method is bottom-up methods, which is not sensitive to the edge information. It sometimes fails to render complete boundary information in very complex scenarios. We hope to mitigate this issue by exploring various forms of edge processing methods in the future.

\section{Conclusion}
In this paper, we proposed a novel bottom-up saliency detection framework
for panoramic images. We first employed a spatial density
 pattern detection method based on dual-stage region growing for the panoramic image to obtain the proposal regions.
Meanwhile, the fixation prediction model was embedded into the framework to locate the focus of attention in images.
Then, the previous saliency information was combined by the maxima normalization to obtain the coarse saliency map. Finally, a geodesic refinement was utilized to get the final saliency map.

Experimental results demonstrated that the proposed saliency detection algorithm provided reliable saliency maps for
panoramic images, and outperformed the recent state-of-the-art saliency
detection methods qualitatively and quantitatively. Moreover,
the proposed algorithm also yielded competitive performance
to the existing saliency detection methods on the conventional
image dataset with common aspect ratios.

Our future research efforts
include the extension of SalPan dataset and the evaluation of algorithm performance on more diverse panoramic scenes.
To encourage future work, we will make the source
codes, experiment data, SalPan dataset for the research community public.

\section*{Acknowledgment}
This work was supported by the grant of National Natural Science Foundation of China(No.U1611461), the grant of Science and Technology Planning Project of Guangdong Province, China(No.2014B090910001) and the grant of Shenzhen PeacockPlan(No.20130408-183003656).
%
%
%
%


%
%

{\small
\bibliographystyle{ieee}
\bibliography{egbib}

\begin{thebibliography}{10}\itemsep=-1pt

\bibitem{Achanta2009}
R.~Achanta, S.~Hemami, F.~Estrada, and S.~Susstrunk.
\newblock Frequency-tuned salient region detection.
\newblock In {\em Computer Vision and Pattern Recognition, 2009. CVPR 2009.
  IEEE Conference on}, pages 1597--1604, 2009.

\bibitem{Alexe2012}
B.~Alexe, T.~Deselaers, and V.~Ferrari.
\newblock Measuring the objectness of image windows.
\newblock {\em IEEE Transactions on Pattern Analysis and Machine Intelligence},
  34(11):2189, 2012.

\bibitem{Borji2012}
A.~Borji.
\newblock Boosting bottom-up and top-down visual features for saliency
  estimation.
\newblock In {\em Computer Vision and Pattern Recognition}, pages 438--445,
  2012.

\bibitem{Borji2015What}
A.~Borji.
\newblock What is a salient object? a dataset and a baseline model for salient
  object detection.
\newblock {\em IEEE Transactions on Image Processing}, 24(2):742, 2015.

\bibitem{Borji2014Salient}
A.~Borji, M.~M. Cheng, H.~Jiang, and J.~Li.
\newblock Salient object detection: A survey.
\newblock {\em Eprint Arxiv}, 16(7):3118, 2014.

\bibitem{Cerf2008}
M.~Cerf, J.~Harel, W.~Einh?user, and C.~Koch.
\newblock Predicting human gaze using low-level saliency combined with face
  detection.
\newblock In {\em International Conference on Neural Information Processing
  Systems}, pages 241--248, 2008.

\bibitem{Cheng2013}
M.~M. Cheng, J.~Warrell, W.~Y. Lin, S.~Zheng, V.~Vineet, and N.~Crook.
\newblock Efficient salient region detection with soft image abstraction.
\newblock In {\em IEEE International Conference on Computer Vision}, pages
  1529--1536, 2013.

\bibitem{fan2017structure}
D.-P. Fan, M.-M. Cheng, Y.~Liu, T.~Li, and A.~Borji.
\newblock Structure-measure: A new way to evaluate foreground maps.
\newblock In {\em Proceedings of the IEEE Conference on Computer Vision and
  Pattern Recognition}, pages 4548--4557, 2017.

\bibitem{Frintrop2010}
S.~Frintrop.
\newblock General object tracking with a component-based target descriptor.
\newblock In {\em IEEE International Conference on Robotics and Automation},
  pages 4531--4536, 2010.

\bibitem{Gao2009}
D.~Gao, S.~Han, and N.~Vasconcelos.
\newblock Discriminant saliency, the detection of suspicious coincidences, and
  applications to visual recognition.
\newblock {\em IEEE Transactions on Pattern Analysis and Machine Intelligence},
  31(6):989--1005, 2009.

\bibitem{Gao2005}
D.~Gao and N.~Vasconcelos.
\newblock Discriminant saliency for visual recognition from cluttered scenes.
\newblock {\em Advances in Neural Information Processing Systems}, 17:481--488,
  2004.

\bibitem{He2016Delving}
K.~He, X.~Zhang, S.~Ren, and J.~Sun.
\newblock Delving deep into rectifiers: Surpassing human-level performance on
  imagenet classification.
\newblock In {\em IEEE International Conference on Computer Vision}, pages
  1026--1034, 2016.

\bibitem{Perazzi2012}
A.~Hornung, Y.~Pritch, P.~Krahenbuhl, and F.~Perazzi.
\newblock Saliency filters: Contrast based filtering for salient region
  detection.
\newblock In {\em IEEE Conference on Computer Vision and Pattern Recognition},
  pages 733--740, 2012.

\bibitem{Hou2007}
X.~Hou and L.~Zhang.
\newblock Saliency detection: A spectral residual approach.
\newblock In {\em IEEE Conference on Computer Vision and Pattern Recognition},
  pages 1--8, 2007.

\bibitem{Hou2009}
X.~Hou and L.~Zhang.
\newblock Dynamic visual attention: Searching for coding length increments.
\newblock In {\em Conference on Neural Information Processing Systems,
  Vancouver, British Columbia, Canada, December}, pages 681--688, 2009.

\bibitem{Itti2004}
L.~Itti.
\newblock Automatic foveation for video compression using a neurobiological
  model of visual attention.
\newblock {\em IEEE Transactions on Image Processing A Publication of the IEEE
  Signal Processing Society}, 13(10):1304, 2004.

\bibitem{Itti1998A}
L.~Itti, C.~Koch, and E.~Niebur.
\newblock A model of saliency-based visual attention for rapid scene analysis.
\newblock {\em IEEE Transactions on Pattern Analysis and Machine Intelligence},
  20(11):1254--1259, 1998.

\bibitem{Jiang2011Automatic}
H.~Jiang, J.~Wang, Z.~Yuan, T.~Liu, N.~Zheng, and S.~Li.
\newblock Automatic salient object segmentation based on context and shape
  prior.
\newblock In {\em British Machine Vision Conference}, 2011.

\bibitem{Kim2014}
J.~S. Kim, J.~Y. Sim, and C.~S. Kim.
\newblock Multiscale saliency detection using random walk with restart.
\newblock {\em IEEE Transactions on Circuits and Systems for Video Technology},
  24(2):198--210, 2014.

\bibitem{Li2015Visual}
G.~Li and Y.~Yu.
\newblock Visual saliency based on multiscale deep features.
\newblock In {\em Computer Vision and Pattern Recognition}, pages 5455--5463,
  2015.

\bibitem{Li2016Deep}
G.~Li and Y.~Yu.
\newblock Deep contrast learning for salient object detection.
\newblock In {\em Computer Vision and Pattern Recognition}, pages 478--487,
  2016.

\bibitem{8265566}
G.~Li and C.~Zhu.
\newblock A three-pathway psychobiological framework of salient object
  detection using stereoscopic technology.
\newblock In {\em 2017 ICCVW}, pages 3008--3014, Oct 2017.

\bibitem{Li2015}
H.~Li, H.~Lu, Z.~Lin, X.~Shen, and B.~Price.
\newblock {\em Inner and Inter Label Propagation: Salient Object Detection in
  the Wild}.
\newblock New Park Pub.,, 2015.

\bibitem{Li2013Saliency}
X.~Li, H.~Lu, L.~Zhang, R.~Xiang, and M.~H. Yang.
\newblock Saliency detection via dense and sparse reconstruction.
\newblock In {\em IEEE International Conference on Computer Vision}, pages
  2976--2983, 2013.

\bibitem{Lin2015}
C.~C. Lin, S.~U. Pankanti, K.~N. Ramamurthy, and A.~Y. Aravkin.
\newblock Adaptive as-natural-as-possible image stitching.
\newblock In {\em IEEE Conference on Computer Vision and Pattern Recognition},
  pages 1155--1163, 2015.

\bibitem{Liu2014Superpixel}
Z.~Liu, X.~Zhang, S.~Luo, and O.~L. Meur.
\newblock Superpixel-based spatiotemporal saliency detection.
\newblock {\em IEEE Transactions on Circuits and Systems for Video Technology},
  24(9):1522--1540, 2014.

\bibitem{Lou2016}
J.~Lou, W.~Zhu, H.~Wang, and M.~Ren.
\newblock Small target detection combining regional stability and saliency in a
  color image.
\newblock {\em Multimedia Tools and Applications}, 76(13):14781--14798, 2017.

\bibitem{Lucchi2012Supervoxel}
A.~Lucchi, K.~Smith, R.~Achanta, G.~Knott, and P.~Fua.
\newblock Supervoxel-based segmentation of mitochondria in em image stacks with
  learned shape features.
\newblock {\em IEEE Transactions on Medical Imaging}, 31(2):474--486, 2012.

\bibitem{Luo2011}
Y.~Luo, J.~Yuan, P.~Xue, and Q.~Tian.
\newblock Saliency density maximization for efficient visual objects discovery.
\newblock {\em IEEE Transactions on Circuits and Systems for Video Technology},
  21(12):1822--1834, 2011.

\bibitem{Miller2009Viewpoint}
D.~R. G.~J. Miller.
\newblock Viewpoint invariant texture description using fractal analysis.
\newblock {\em International Journal of Computer Vision}, 83(1):85--100, 2009.

\bibitem{Oliva2003}
A.~Oliva, A.~Torralba, M.~S. Castelhano, and J.~M. Henderson.
\newblock Top-down control of visual attention in object detection.
\newblock In {\em International Conference on Image Processing, 2003. ICIP
  2003. Proceedings}, pages I--253--6 vol.1, 2003.

\bibitem{Qin2015Saliency}
Y.~Qin, H.~Lu, Y.~Xu, and H.~Wang.
\newblock Saliency detection via cellular automata.
\newblock In {\em Computer Vision and Pattern Recognition}, pages 110--119,
  2015.

\bibitem{Margolin2013}
M.~Ran, A.~Tal, and L.~Zelnikmanor.
\newblock What makes a patch distinct?
\newblock In {\em IEEE Conference on Computer Vision and Pattern Recognition},
  pages 1139--1146, 2013.

\bibitem{Rosenholtz2011}
R.~Rosenholtz, A.~Dorai, and R.~Freeman.
\newblock Do predictions of visual perception aid design?
\newblock {\em Acm Transactions on Applied Perception}, 8(2):1--20, 2011.

\bibitem{Rother2004}
C.~Rother, V.~Kolmogorov, and A.~Blake.
\newblock "grabcut": interactive foreground extraction using iterated graph
  cuts.
\newblock In {\em ACM SIGGRAPH}, pages 309--314, 2004.

\bibitem{Shi2016Hierarchical}
J.~Shi, Q.~Yan, X.~Li, and J.~Jia.
\newblock Hierarchical image saliency detection on extended cssd.
\newblock {\em IEEE Transactions on Pattern Analysis and Machine Intelligence},
  38(4):717--729, 2016.

\bibitem{Shi2016}
J.~Shi, Q.~Yan, X.~Li, and J.~Jia.
\newblock Hierarchical image saliency detection on extended cssd.
\newblock {\em IEEE Transactions on Pattern Analysis and Machine Intelligence},
  38(4):717, 2016.

\bibitem{Siagian2009}
C.~Siagian and L.~Itti.
\newblock Biologically inspired mobile robot vision localization.
\newblock {\em IEEE Transactions on Robotics}, 25(4):861--873, 2009.

\bibitem{Tenenbaum2000}
J.~B. Tenenbaum, S.~V. De, and J.~C. Langford.
\newblock A global geometric framework for nonlinear dimensionality reduction.
\newblock {\em Science}, 290(5500):2319, 2000.

\bibitem{Judd2009}
J.~Tilke, K.~Ehinger, F.~Durand, and A.~Torralba.
\newblock Learning to predict where humans look.
\newblock 30(2):2106--2113, 2009.

\bibitem{Tong2014Saliency}
N.~Tong, H.~Lu, L.~Zhang, and R.~Xiang.
\newblock Saliency detection with multi-scale superpixels.
\newblock {\em IEEE Signal Processing Letters}, 21(9):1035--1039, 2014.

\bibitem{Wang2017}
J.~Wang, H.~Jiang, Z.~Yuan, M.-M. Cheng, X.~Hu, and N.~Zheng.
\newblock Salient object detection: A discriminative regional feature
  integration approach.
\newblock {\em International Journal of Computer Vision}, 123(2):251--268, Jun
  2017.

\bibitem{wang2017learning}
L.~Wang, H.~Lu, Y.~Wang, M.~Feng, D.~Wang, B.~Yin, and X.~Ruan.
\newblock Learning to detect salient objects with image-level supervision.
\newblock In {\em Proceedings of the IEEE Conference on Computer Vision and
  Pattern Recognition}, pages 136--145, 2017.

\bibitem{Wang2016Saliency}
L.~Wang, L.~Wang, H.~Lu, P.~Zhang, and R.~Xiang.
\newblock Saliency detection with recurrent fully convolutional networks.
\newblock In {\em European Conference on Computer Vision}, pages 825--841,
  2016.

\bibitem{wang2016kernelized}
T.~Wang, L.~Zhang, H.~Lu, C.~Sun, and J.~Qi.
\newblock Kernelized subspace ranking for saliency detection.
\newblock In {\em European Conference on Computer Vision}, pages 450--466.
  Springer, 2016.

\bibitem{Wei2012}
Y.~Wei, F.~Wen, W.~Zhu, and J.~Sun.
\newblock {\em Geodesic Saliency Using Background Priors}.
\newblock Springer Berlin Heidelberg, 2012.

\bibitem{Yang2013}
C.~Yang, L.~Zhang, and H.~Lu.
\newblock Graph-regularized saliency detection with convex-hull-based center
  prior.
\newblock {\em IEEE Signal Processing Letters}, 20(7):637--640, 2013.

\bibitem{Yang2013Saliency}
C.~Yang, L.~Zhang, H.~Lu, X.~Ruan, and M.~H. Yang.
\newblock Saliency detection via graph-based manifold ranking.
\newblock In {\em IEEE Conference on Computer Vision and Pattern Recognition},
  pages 3166--3173, 2013.

\bibitem{Yang2012}
J.~Yang.
\newblock Top-down visual saliency via joint crf and dictionary learning.
\newblock In {\em Computer Vision and Pattern Recognition}, pages 2296--2303,
  2012.

\bibitem{zhang2017amulet}
P.~Zhang, D.~Wang, H.~Lu, H.~Wang, and X.~Ruan.
\newblock Amulet: Aggregating multi-level convolutional features for salient
  object detection.
\newblock In {\em Proceedings of the IEEE Conference on Computer Vision and
  Pattern Recognition}, pages 202--211, 2017.

\bibitem{Zhao2015Saliency}
R.~Zhao, W.~Ouyang, H.~Li, and X.~Wang.
\newblock Saliency detection by multi-context deep learning.
\newblock In {\em Computer Vision and Pattern Recognition}, pages 1265--1274,
  2015.

\bibitem{Zhou2013Ranking}
Zhou, Dengyong, Weston, Jason, Gretton, Arthur, Bousquet, Olivier, Sch?lkopf,
  and Bernhard.
\newblock Ranking on data manifolds.
\newblock {\em Advances in Neural Information Processing Systems}, pages
  169--176, 2013.

\bibitem{PDNet2018}
C.~Zhu, X.~Cai, H.~Kan, L.~Thomas.H, and L.~Ge.
\newblock Pdnet: Prior-model guided depth-enhanced network for salient object
  detection.
\newblock In {\em 2018 International Conference on Multimedia and Expo}, 2018.

\bibitem{Zhu2018An}
C.~Zhu, K.~Huang, and G.~Li.
\newblock An innovative saliency guided roi selection model for panoramic
  images compression.
\newblock In {\em DCC}, pages 438--438, 2018.

\bibitem{Zhu2018}
C.~Zhu and G.~Li.
\newblock A multilayer backpropagation saliency detection algorithm and its
  applications.
\newblock {\em Multimedia Tools and Applications}, Mar 2018.

\bibitem{10.1007/978-3-319-64698-5_2}
C.~Zhu, G.~Li, X.~Guo, W.~Wang, and R.~Wang.
\newblock A multilayer backpropagation saliency detection algorithm based on
  depth mining.
\newblock In {\em CAIP}, pages 14--23, 2017.

\bibitem{8265388}
C.~Zhu, G.~Li, W.~Wang, and R.~Wang.
\newblock An innovative salient object detection using center-dark channel
  prior.
\newblock In {\em 2017 ICCVW}, pages 1509--1515, Oct 2017.

\bibitem{ZhuICCV2017}
C.~Zhu, G.~Li, W.~Wang, and R.~Wang.
\newblock An innovative salient object detection using center-dark channel
  prior.
\newblock In {\em The IEEE International Conference on Computer Vision (ICCV)},
  Oct 2017.

\bibitem{7966712}
C.~Zhu, G.~Li, W.~Wang, and R.~Wang.
\newblock Salient object detection with complex scene based on cognitive
  neuroscience.
\newblock In {\em 2017 IEEE Third International Conference on Multimedia Big
  Data (BigMM)}, pages 33--37, April 2017.

\bibitem{8265548}
C.~Zhu, T.~H. Li, and G.~Li.
\newblock Towards automatic wild animal detection in low quality camera-trap
  images using two-channeled perceiving residual pyramid networks.
\newblock In {\em 2017 ICCVW}, pages 2860--2864, Oct 2017.

\bibitem{Zhu2014}
W.~Zhu, S.~Liang, Y.~Wei, and J.~Sun.
\newblock Saliency optimization from robust background detection.
\newblock In {\em Computer Vision and Pattern Recognition}, pages 2814--2821,
  2014.

\end{thebibliography}
}

\end{document}